\documentclass[letterpaper]{article} % DO NOT CHANGE THIS
\usepackage{aaai25}  % DO NOT CHANGE THIS
\usepackage{times}  % DO NOT CHANGE THIS
\usepackage{helvet}  % DO NOT CHANGE THIS
\usepackage{courier}  % DO NOT CHANGE THIS
\usepackage[hyphens]{url}  % DO NOT CHANGE THIS
\usepackage{graphicx} % DO NOT CHANGE THIS
\usepackage{natbib}  % DO NOT CHANGE THIS AND DO NOT ADD ANY OPTIONS TO IT
\usepackage{caption} % DO NOT CHANGE THIS AND DO NOT ADD ANY OPTIONS TO IT
\usepackage{algorithm}
\usepackage{algorithmic}

\usepackage{newfloat}
\usepackage{listings}
\DeclareCaptionStyle{ruled}{labelfont=normalfont,labelsep=colon,strut=off} % DO NOT CHANGE THIS
\floatstyle{ruled}
\newfloat{listing}{tb}{lst}{}
\floatname{listing}{Listing}
\usepackage{bibentry}
\usepackage{latexsym}
\usepackage{amssymb}
\usepackage{amsmath}
\usepackage{amsthm}
\usepackage{booktabs}
\usepackage{enumitem}
\usepackage{graphicx}
\usepackage{color}
\usepackage{verbatim}

\usepackage{microtype}
\usepackage{subfigure}
\usepackage{array}
\usepackage{blindtext}
\usepackage{multirow}
\usepackage{mwe}
\usepackage{optidef}
\usepackage{lipsum} 
\RequirePackage{algorithm}
\RequirePackage{algorithmic}
\RequirePackage{natbib}
\RequirePackage{eso-pic} % used by \AddToShipoutPicture
\RequirePackage{url}
\usepackage{xr}
\title{Leveraging Constraint Violation Signals For Action-Constrained \\Reinforcement Learning}
\author{
      Janaka Chathuranga Brahmanage, Jiajing Ling, Akshat Kumar \\
}
\affiliations{
School of Computing and Information Systems, Singapore Management University.\\
\{janakat.2022, jjling.2018\}@phdcs.smu.edu.sg, akshatkumar@smu.edu.sg

}

\newcommand{\our}{$\mathbb{CV}$-Flows}
\newcommand{\ourcv}{$\mathbb{CV}$}
\DeclareMathOperator {\cv}{{CV}}
\DeclareMathOperator {\kl}{KL}

\newcommand{\mc}{\mathcal}

\newcommand{\la}{\leftarrow}

\newcommand{\squishlist}{
	\begin{list}{$\bullet$}
		{ \setlength{\itemsep}{0pt}
			\setlength{\parsep}{3pt}
			\setlength{\topsep}{3pt}
			\setlength{\partopsep}{0pt}
			\setlength{\leftmargin}{1.5em}
			\setlength{\labelwidth}{1em}
			\setlength{\labelsep}{0.5em} } }
\newcommand{\squishlisttwo}{
	\begin{list}{$\bullet$}
		{ \setlength{\itemsep}{0pt}
			\setlength{\parsep}{3pt}
			\setlength{\topsep}{3pt}
			\setlength{\partopsep}{0pt}
			\setlength{\leftmargin}{1.5em}
			\setlength{\labelwidth}{1em}
			\setlength{\labelsep}{0.5em} } }
\newcommand{\squishend}{
	\end{list}  }
\newcommand{\red}[1]{#1}
\newcommand{\blue}[1]{#1}
\newtheorem{prop}{Proposition}

\DeclareMathOperator*{\E}{\mathop{\mathbb{E}}}
\DeclareMathOperator*{\argmin}{arg\,min}

\begin{document}

\maketitle

\begin{abstract}
In many RL applications, ensuring an agent's actions adhere to constraints is crucial for safety. Most previous methods in Action-Constrained Reinforcement Learning (ACRL) employ a projection layer after the policy network to correct the action. However projection-based methods suffer from issues like the zero gradient problem and higher runtime due to the usage of optimization solvers. Recently methods were proposed to train generative models to learn a differentiable mapping between latent variables and feasible actions to address this issue. However, generative models require training using samples from the constrained action space, which itself is challenging. To address such limitations, \textit{first}, we define a target distribution for feasible actions based on constraint violation signals, and train normalizing flows by minimizing the KL divergence between an approximated distribution over feasible actions and the target. This eliminates the need to generate feasible action samples, greatly simplifying the flow model learning. \textit{Second}, we integrate the learned flow model with existing deep RL methods, which restrict it to exploring only the feasible action space. \textit{Third}, we extend our approach beyond ACRL to handle state-wise constraints by learning the constraint violation signal from the environment. Empirically, our approach has significantly fewer constraint violations while achieving similar or better quality in several control tasks than previous best methods.
\end{abstract}

\begin{links}
    \link{Code}{https://github.com/rlr-smu/cv-flow}
\end{links}

\section{Introduction}

Reinforcement learning has been successfully applied to solve a variety of problems, ranging from mastering Atari games~\cite{mnih2015human,vanhasseltDeepReinforcementLearning2015} to controlling robotics~\cite{pham2018optlayer, thananjeyanRecoveryRLSafe2021} and fortifying system security~\cite{ adawadkarCybersecurityReinforcementLearning2022,khouryHybridGameTheory2020}, among others. {In many real-world applications, agents have to take actions within a feasible action space defined by some constraints at every RL step. This scenario falls under the domain of action-constrained reinforcement learning (ACRL)~\cite{brahmanage2023flowpg,Kasaura23}. In ACRL, action constraints typically take an analytical form based on state and action features (e.g., $\sum_{i=1}^2|s_ia_i|<= 1$ with $s_i, a_i$ as features) rather than being expressed using a predefined cost function. 
% mnihPlayingAtariDeep2013 - Alternative citation

\blue{
Representative applications of ACRL include robotic control~\cite{Kasaura23,lin2021escaping,pham2018optlayer} and resource allocation in supply-demand matching~\cite{Bha19}, where kinematic limitations and resource restrictions are effectively modeled using action constraints derived from system specifications.
When action constraints are not directly available, they can be approximated using state-based cost functions or offline datasets of valid and invalid trajectories. In such cases, action constraints can be inferred from environmental data~\cite{dalal2018safe} or via inverse constraint RL~\cite{MalikInverseConstrainedReinforcement2021}.
}

Since action constraints with a closed form can be relatively easy to evaluate for each action, one natural approach is to use a \textit{ projection} to generate feasible actions that satisfy the constraints, which involves solving a math program~\cite{amos2017optnet,Bha19,dalal2018safe,lin2021escaping,pham2018optlayer}. The projection-based approach can either result in a zero gradient problem~\cite{lin2021escaping} or an expensive overhead due to solving an optimization program~\cite{brahmanage2023flowpg} or both. When dealing with non-convex action constraints or large action spaces, there is a significant reduction in training speed, which we also validate empirically.

Recently,~\cite{brahmanage2023flowpg} learn a smooth and invertible mapping between a simple latent space and the feasible action space given states using conditional normalizing flows~\cite{dinh2016density} and integrate it with Deep Deterministic Policy Gradient~(DDPG)~\cite{lillicrap2015continuous}. This approach effectively avoids the zero gradient problem common in standard projection-based methods as there is no coupling between the policy network and projection layer anymore. However, a key drawback is the necessity to generate feasible actions in advance for training the flow, which is challenging for complex constraints since specialized methods such as Hamiltonian Monte-Carlo~(HMC), decision diagrams are required to sample from the feasible action space~\cite{brahmanage2023flowpg}.

In addition to ACRL, generative models are also used for improving exploration and utilizing off-policy data in RL. 
In the context of enhancing exploration~\cite{mazoure2020leveraging,ward2019improving}, generative models are used as a policy for optimizing the maximum entropy objective of Soft Actor Critic~(SAC)~\cite{haarnojaSoftActorCriticOffPolicy2018}.
Also, ~\cite{fujimoto2019offpolicy,zhang2023apac} utilize off-policy data to train a normalizing flow to improve the exploration. Recent work ~\cite{changyuGenerativeModellingStochastic} uses Argmax Flow (a type of normalizing flow) with RL to deal with discrete action spaces.

\paragraph{Contributions}
Our main contributions are as follows. 

\textit{First}, we learn an invertible, differentiable mapping from a simple base distribution (e.g., Gaussian) of the normalized flow model to the feasible action space. Standard methods for flow model training require generating data (feasible environment action samples) using specialized methods such as HMC and decision diagrams~\cite{dinh2016density,brahmanage2023flowpg}. {Our method instead directly trains the flow by minimizing the KL divergence between a flow model based distribution and a target density over the feasible action space defined using the constraint violation signal.} This avoids the costly step of generating samples from the feasible action space. 

\textit{Second}, we integrate such a trained flow model with SAC~\cite{haarnojaSoftActorCriticOffPolicy2018}. 
We also present an analytical method for computing entropy exclusively over the feasible action space, which offers significant advantages over the naive implementation of SAC for ACRL.
\blue{This method can avoid the zero-gradient problem, as a properly calibrated flow model ensures a one-to-one mapping from the support of a simple distribution (e.g., Gaussian) to the feasible action space without requiring an optimization-based projection layer.}

\textit{Third}, we present a methodology to extend our approach beyond ACRL to address state-wise constraints (non-explicit constraints). We achieve this by learning the constraint violation signal through interaction with the environment.

\textit{Finally}, we train the flow model on a variety of action constraints and state-wise constraints used in the ACRL literature, leading to an accurate approximation of the feasible action space. Our approach results in fewer constraint violations ({$>$10x} for a number of benchmarks) while achieving similar or better solution qualities on a variety of continuous control tasks than the previous best methods~\cite{brahmanage2023flowpg,Kasaura23,lin2021escaping}.

\section{Preliminaries} \label{sec:preliminaries}
\subsection{Action-Constrained MDP} 
An action-constrained Markov Decision Process (MDP) is a standard MDP augmented with explicit action constraints. An MDP is defined using a tuple $\langle S, A, p, r, \gamma, b_0 \rangle$, where $S$ is the set of possible states, $A$ is the unconstrained action space--a set of possible actions that an agent can take. The $p(s'|s, a)$ is the probability of leading to state $s'$ after taking action $a$ in state $s$; $r(s, a)$ is the immediate reward received after taking action $a$ in state $s$; $\gamma \in [0, 1)$ is the discount factor, and $b_0$ is the initial state distribution. 
Given a state $s \in S$, we define a feasible action space $\mathcal{C}(s) \subseteq A$ using $m$ inequality and $n$ equality constraints as:
% \vskip -1pt 
{
\small
\begin{align}\label{eq:action_constraints}
\!\!\mathcal{C}(s) \!=\! \{a | a \!\in\! A, g_i(a, s) \!\leq\! 0, h_j(a, s)\!=\!0, i \!=\! 1\!:\!m, j \!=\! 1\!:\!n \}
\end{align}
}
where \(g_i\) and \(h_j\) are arbitrary functions of state and action used to define inequality and equality constraints.
We assume continuous state and action spaces. Let $\mu_\phi$ denote the policy parameterized by $\phi$. For a stochastic policy, we use $\mu_\phi(a|s)$ to denote the probability of taking action $a$ in state $s$. 
In RL setting transition and reward functions are not known. The agent learns a policy by interacting with the environment and using collected experiences $(s_t, a_t, r_t, s_{t+1})$.  
The goal is to find a policy that maximizes the expected discounted total reward while ensuring that all chosen actions are taken from the feasible action space:  
% \vskip -1pt
{
\small
\begin{maxi}|s|
{\phi}{
J(\mu_\phi) = \mathbb{E}_{ \mu_\phi, s_0 \sim b_0}[\sum_{t=0}^\infty \gamma^t r(s_t, a_t)]
} 
{\label{eq:acrl}}{}
\addConstraint{a_t \in \mathcal{C}(s_t)\ \quad \forall t}
\end{maxi}  
}

In ACRL, the environment simulator only accepts feasible actions, as assumed in previous work~\cite{lin2021escaping,Kasaura23}. Any infeasible actions will lead to the termination of the simulator. That is why a projection is typically used to ensure that any infeasible actions are mapped into the feasible action space.

\subsection{\textit{State-Constrained} MDP} \label{sec:state-wise}
In some real-world problems, constraints only involve state features. For example, in safe RL, 
when an agent takes an action $a$ in a state $s$, it must ensure that the next state $s'$ is safe~\cite{dalal2018safe, zhaoStatewiseSafeReinforcement2023}. These are referred to as state-wise constraints. Mathematically, they can be defined as $c_i(s) \leq 0, i = 1\ldots k $. Here $c_i$ are state based cost functions. As shown in~\cite{dalal2018safe}, each state-wise constraint can be transformed into an action constraint by approximating the constraint violations in the next state using a neural network $w_i$ as follows.
\begin{equation}\label{eq:state-wise-constraints}
  c_i(s_{t+1})\approx c_i(s_{t})+w_i(s_t)^Ta_t \leq 0,\quad i = 1\ldots k  
\end{equation}

As a pretraining step, we run a random policy to collect experience $(s_t, a_t, r_t, s_{t+1})$, and the associated costs $c(s_t)$ and $c(s_{t+1})$ for the current and next states. 
We then train the neural network $w_i$ by minimize the Mean Squared Error (MSE) loss between the predicted cost $c_i(s_t) + w_i(s_t)^T a_t$ and the observed cost $c_i(s_{t+1})$. 
Once trained, this model defines a feasible action space $\mc{C}(s) = \{a | a \in A, c_i(s) + w_i(s)^T a \leq 0 ; \forall i\}$.

\subsection{Existing Approaches to Solving ACRL}

\begin{figure}[tb]
    \centering	
    \includegraphics[width=\linewidth]{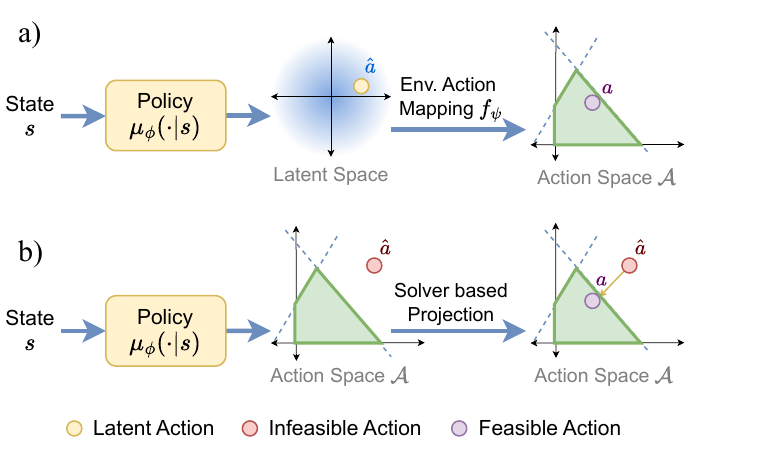}
    % \vspace{-0.5cm}
    \caption{Two approaches to integrate action constraints with RL: (a) Mapping-based approach and (b) Projection-based approach.}
    \label{fig:type_acrl}
    % \vspace{-0.1cm}
\end{figure}

As discussed in the introduction, there are two popular approaches to solving ACRL. 
Figure~\ref{fig:type_acrl}(b) illustrates the projection-based approach. In this method, the policy network $\mu$ parameterized by $\phi$ receives a state $s$ and outputs an action $\hat{a}$. When $\hat{a}$ is infeasible—typically the case at the beginning of the training stage—a quadratic programming (QP) solver is used to project $\hat{a}$ into the feasible action space, resulting in a feasible environment action $a$. 
Figure~\ref{fig:type_acrl}(a) shows the mapping-based approach, where the policy network $\mu_{\phi}$ generates $\hat{a}$ in a latent space rather than the environment action space. 
This latent action $\hat{a}$ is then converted into an environment action using a mapping function $f$ parameterized by $\psi$, such that $a = f_\psi(\hat{a})$. 

\blue{We use a mapping-based approach since it offers several advantages over a projection layer. Unlike a projection layer, which suffers from the zero-gradient problem—where multiple infeasible actions map to a single action, limiting the agent's ability to learn effectively—and distorts probability density by concentrating actions in bordering regions, the mapping function avoids these issues and ensures that the agent can explore the feasible action space evenly. In our work, we use normalizing flows as the mapping function $f_\psi$.}

\subsection{Normalizing Flows} 
A normalizing flow model is a type of generative model. 
It transforms a simple \textit{base distribution} such as a Gaussian into a more complex distribution through an invertible transformation function~\cite{rezende2015variational,dinh2016density}. 

In the context of solving ACRL, normalizing flows are used to map the base distribution into a feasible environment action density distribution. Given a state $s$, a sample $\hat{a}$ from the base distribution $\hat{q}$, and state-conditioned normalizing flows $f$ parameterized by $\psi$, we obtain a feasible environment action $a$ as $a = f_{\psi}(\hat{a}, s)$.
\blue{Let $q(a|s)$ denote the probability of obtaining $a$ through normalizing flows. Since $f$ is bijective,  the log probability can be computed using the change of variables theorem as follows:}
\begin{align} \label{eq:conditional_nvp_prob}
\log q(a|s) = \log \frac{{\hat{q}(\hat{a})}}{|\det J_{f_\psi}(\hat{a};s)|}
\end{align}
\blue{where $|\det J_{f_\psi}(\hat{a};s)|$ is the determinant of the Jacobian of $f_\psi$, which accounts for the change in volume when transforming the base distribution to the feasible action density distribution~\cite{nielsenSurVAEFlowsSurjections2020}.} 

When a training dataset of feasible state-action pairs is available, the function $f$ is learned by maximizing the log-likelihood of the data~\cite{dinh2016density}. 
Conversely, when feasible state-action pairs are not provided but \blue{the feasible environment action distribution conditioned on state} is known, $f$ can be learned by minimizing the reverse Kullback-Leibler (KL) divergence between the true target distribution denoted by $p(a|s)$ and $q(a|s)$~\cite{papamakarios2021normalizing}. 
In ACRL, we want the model to generate all feasible actions with equal likelihood while avoiding any infeasible actions. Therefore, we define \(p(a|s)\) as a uniform distribution over feasible environment actions, assigning zero probability to infeasible actions.
Given that the reverse-KL loss for a given state $s$ is:
% \vskip -1pt
{
\small
\begin{align} \label{eq:reverse_kl}
\mathcal{L}(s) &= \kl(q||p)= \E_{\substack{\hat{a} \sim \hat{q}(\hat{a})\\a=f_\psi(\hat{a},s)}} [\log q(a|s)- \log p(a|s)]\nonumber \\
&= \E_{\hat{a} \sim q(\hat{a})} \left[\log \frac{\hat{q}(\hat{a})}{|\det J_{f_\psi}(\hat{a};s)|} - \log p(f_\psi(\hat{a},s)|s) \right]
\end{align}
}

\section{Methodology}
\subsection{Overview}

\begin{figure}[tb]
    \centering	
    \includegraphics[width=\linewidth]{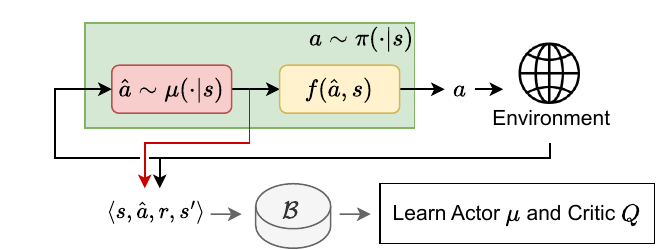}
    % \vspace{-0.5cm}
    \caption{
    % Flow Model integration with the SAC Policy. % TODO Add description of the figure here.
    Flow Model integration with the SAC Policy: \(\mu\) represents the original policy network, \(\hat{a}\) is the latent action, and \(f\) is the mapping function that maps the latent action into a feasible environment action \(a\). \(\pi\) represents the combined policy. The latent action \(\hat{a}\) is stored in the replay buffer to train both the \(\mu\) and critic networks.
    }
    \label{fig:flow_agent}
\end{figure}

Our architecture, as shown in Figure~\ref{fig:flow_agent}, consists of two components: the policy \(\mu_\phi\) and the mapping function \(f_\psi\). 
We refer to the combination of these two as the combined policy \(\pi_{\phi,\psi}\). The training process involves two primary steps. 
First, we train the mapping function \(f_\psi\), a normalizing flow. 
Unlike previous approaches, we do not assume the availability of a dataset of feasible environment actions, so we utilize the reverse KL divergence Eq.~\eqref{eq:reverse_kl} to train the flow model. 
Second, we train the reinforcement learning agent. 
\blue{In contrast to previous work~\cite{brahmanage2023flowpg}, we use the latent action $\hat{a}$ instead of environment action $\hat{a}$ to train both the actor and critic networks. }
This approach offers a computational advantage, as we do not need to backpropagate through the flow model during RL training. 
In Sections~\ref{sec:approx_const} and~\ref{sec:sac_cv_flow}, we discuss these steps in detail.

\subsection{Feasible Action Mapping Using \our}\label{sec:approx_const}
As noted in Section~\ref{sec:preliminaries}, a standard approach for training normalizing flows as a mapping function is to maximize the log-likelihood over the training dataset (feasible state-action pairs in our setting). However, a key challenge lies in generating a sufficient number of feasible samples from the constrained action space. This often requires techniques like rejection sampling or advanced methods such as Markov-Chain Monte Carlo (MCMC)~\cite{mcmc12}, which can be computationally intractable and sample inefficient with high rejection rate for complex constraints as shown in~\cite{brahmanage2023flowpg}.

To eliminate the need to generate feasible state-action samples to train normalizing flows, we propose a novel approach, namely Constraint Violation-Flows (\our) to train the normalizing flows. We first define a true target distribution $p(a|s)$ over feasible state-action samples using constraint violation signals. Then we train the flow by minimizing the reverse KL divergence between an approximated distribution of feasible samples ($q(a|s)$ in Eq.~\eqref{eq:conditional_nvp_prob}) 
and the true target distribution, as shown in Eq.~\eqref{eq:reverse_kl}. 

\noindent \textbf{Defining the true target distribution:}
We consider the constrained action space defined by Eq.~\eqref{eq:action_constraints}. 
We first define the magnitude of constraint violations ($\cv$) as follows:
% \vskip -1pt
{
\small
\begin{align*}
{\cv(a, s)} \!=\! \sum_{i=1}^m \max(g_i(a, s), 0) \!+\! {\sum_{j=1}^n \max(|~|h_j(a, s)|- \epsilon|, 0)}
\end{align*}\label{eq:cv-signal}
}
\noindent where $\epsilon$ is an error margin for equality constraints. The function {$\cv(a, s)$} evaluates to zero if and only if an action $a$ is within the constrained action space, increasing as it moves away from this feasible region. Then we define a non-negative measure over this region as follows:
\begin{equation}\label{eq:cv_prior}
    \tilde{p}(a|s) = e^{-\lambda\cv(a,s)}
\end{equation}
This ensures that $\tilde{p}(a|s)$ remains positive in the absence of constraint violations and decreases exponentially with increasing $\cv(a,s)$. The rate parameter $\lambda$ of the exponential distribution determines the steepness of the probability decrease as constraint violations increase.
\blue{A larger $\lambda$ is preferred to boost the loss for even small constraint violations. It was set to 1000 for all the experiments.}

Lastly, to obtain a valid probability distribution given $s$, we normalize the non-negative measure by dividing a constant $M(s) = \int_{-\infty}^\infty \tilde{p}(a|s) da$. Thus, we have,
\begin{equation} \label{eq:target_distribution}
     p(a|s) = \frac{\tilde{p}(a|s)}{M(s)} 
\end{equation}
We note that this target distribution is a uniform distribution over feasible environment actions given $s$ since $\cv(a,s)$ is the same for every single feasible environment action. {This is also desirable as it implies that after flow training, $q(a|s)$ will be able to generate most feasible environment actions.}
The probability for infeasible actions will be very close to zero.

\noindent \textbf{Training the flow model:} To train the normalizing flow, we first need to sample a sufficient number of $\hat{a} \sim \hat{q}(\hat{a})$ from a base distribution $\hat{q}$ (e.g., standard Gaussian). Sampling $\hat{a}$ from the standard Gaussian is much easier than sampling feasible environment actions from $p(a|s)$. Substituting the true target distribution Eq.~\eqref{eq:target_distribution} in Eq.~\eqref{eq:reverse_kl}, the loss function (for a state $s$) is: 
{
\begin{align} \label{eq:reverse_kl_tilde}
\mc{L}&(s)\!=\!\E_{\hat{a} \sim \hat{q}(\hat{a})}\left[ \log \frac{\hat{q}(\hat{a})}{|\det J_{f_\psi}(\hat{a};s)|}\!-\!\log \frac{\tilde{p}( f_\psi(\hat{a},s)|s)}{M(s)}\right] \nonumber \\
&\!\!\!\!=\!\!\!\!\!\E_{\hat{a} \sim \hat{q}(\hat{a})}\left[ \log \frac{\hat{q}(\hat{a})}{|\det J_{f_\psi}(\hat{a};s)|}\!-\!\log \frac{e^{-\lambda\cv(f_\psi(\hat{a},s),s)}}{M(s)}\right]
\end{align}
}

\noindent Since both $\hat{q}(\hat{a})$ and $M(s)$ are independent of $\psi$, they can be excluded. Then, given a probability distribution over the state space {$p_S(s)$}, the final loss function can be written as:
% \vskip -1pt
{
\small
\begin{equation}
J^f(\psi) =\!\!\!\E_{s\sim  p_S, \hat{a} \sim \hat{q}} [\lambda \cv(f_\psi(\hat{a},s), s)-\log {|\det J_{f_\psi}(\hat{a};s)|}]
\end{equation}
}

We note that the distribution $p_S(s)$ can be uniform if the state space is bounded; otherwise, it can be Gaussian. 
In the case of a Gaussian distribution, the mean and standard deviation can be determined by running an unconstrained agent in the environment. 
Alternatively, state samples can be collected directly from the environment by operating the agent under a different policy. The pseudo-code of our proposed approach to training the \our~is provided in Algorithm~\ref{alg:ud_flow}. 

\noindent \textbf{Training the flow model for state-wise constraints:} For state-wise constraints we do not have access to the analytical form of the action constraint. Instead, we only have state constraints $c_i(s)$ in the form of Eq.~\eqref{eq:state-wise-constraints}. We follow the linear approximation model discussed in Section~\ref{sec:state-wise}, which results in CV function in Eq.~\eqref{eq:cv-signal-state-wise}:
{
\begin{align}\label{eq:cv-signal-state-wise}
{\cv(a, s)} \!=\! \sum_{i=1}^k \max(c_i(s)+w_i(s)^Ta, 0) 
\end{align}
}

\noindent In Eq.~\eqref{eq:cv-signal-state-wise}, $w_i$ is learned through a pretraining process using the transitions collected by running a random agent~\cite{dalal2018safe} in the environment, as discussed in Section~\ref{sec:state-wise}. Subsequently, we use this CV signal to train the \our, as discussed in the previous section.

\begin{algorithm}[tb]
\caption{\our \ Pretraining Algorithm}\label{alg:ud_flow}
\begin{algorithmic}[1]
\STATE Initialize normalizing flow $f_\psi$, with random weights $\psi$
\FOR{$epoch=1...N$}
\STATE $\hat{a} \sim \hat{q}(\hat{a})$ \COMMENT{Batch sample $\hat{a}$ from the base dist.}
\STATE $s \sim p_S(s)$ \COMMENT{Batch sample $s$ from state space}

\STATE $\psi \la \psi - \lambda_f \hat\nabla_\psi J^f(\psi)$ \COMMENT{{Update flow model}}
\ENDFOR
\end{algorithmic}
\end{algorithm}

\noindent \textbf{Benefits of using \our:}
% The benefits of \our~are two-fold. First, it can be easily integrated with maximum entropy deep RL methods, such as SAC (details will be discussed in the next section). 
% \blue{Assuming a Gaussian distribution as the base, the flow model ensures that any sample drawn from the policy is mapped approximately to the feasible region in a differentiable manner.}
% This mapping does not involve any optimization solvers, unlike other projection-based methods, thereby avoiding the zero-gradient problem in RL policy training. {Second, normalizing flows are advantageous in our setting over other generative models such as GANs and VAEs as in maximum entropy RL, because we need to compute the log probability of generated feasible environment actions. It is easy to get from the flow model because of bijectivity, and not available in GANs or VAEs easily. Additionally, it has been shown in the prior work that flow based models provided better accuracy and recall for mapping to the feasible environment action space~\cite{brahmanage2023flowpg}}.
The benefits of \our~are twofold. First, it integrates seamlessly with maximum entropy deep RL methods like SAC (details in the next section). Assuming a Gaussian base, the flow model maps policy samples to the feasible region differentiably, avoiding optimization solvers and the zero-gradient issue in RL policy training. Second, normalizing flows outperform GANs and VAEs in maximum entropy RL by enabling efficient log-probability computation due to bijectivity. Prior work also shows flow models offer better accuracy and recall for mapping to feasible action spaces~\cite{brahmanage2023flowpg}.

\subsection{Integrating RL Agent With the \our}\label{sec:sac_cv_flow}

As shown in the previous section, the normalizing flow model maps the base distribution to the feasible action space. Here, we demonstrate its integration with the existing RL algorithm SAC~\cite{haarnojaSoftActorCriticOffPolicy2018}. 
We do not change the training loop of the base algorithm. It collects rollout from the environment and save $\langle s, \hat{a}, r, s' \rangle$ are stored in a replay buffer $\mathcal{B}$, which will be used to update the policy network $\mu_\phi$ and the critic network $Q_\theta$ where $\theta$ represents the parameters. But we store the latent action $\hat{a}$ in the replay buffer instead of the final action $a$, because we want to train both critic and policy networks using the latent action to avoid backpropagation through the flow model.
Additionally, during this step, we keep the flow model parameters \(\psi\) fixed.
The mapping function may not be fully accurate, so action \(a\) can violate constraints. To ensure feasibility, we add a projection step to map \(a\) into the feasible region if needed.

The base policy \(\mu_\phi\) produces the latent action \(\hat{a}\), which is mapped to the feasible environment action \(a\) by \(f_\psi\).
Let us call the composition of base policy \(\mu_\phi\) and the normalizing flow \(f_\psi\) that generates a feasible environment action as \(\pi_{\phi, \psi}\).
When writing the objectives for SAC~\cite{haarnojaSoftActorCriticOffPolicy2018} based on this combined policy, we get the following objectives for critic and policy update. 
% \vskip -1pt
{
\small
\begin{align}\label{eq:sac_policy_loss_original}
  J^{\pi}(\phi)= -\E_{s\sim \mathcal{B}, a\sim \pi(\cdot|s)}\big[Q^{\pi(s,a)}(s,a)-\alpha\log \pi(a|s)\big]
\end{align}
}
{
\small
\begin{align}\label{eq:sac_critic_loss_original}
J^{Q}(\theta)=&\E_{\substack{(s,a,r,s')\sim \mc{B}, a'\sim \pi(\cdot|s')}}[(Q_{\theta}(s,a)  \\ &-(r+\gamma (Q_{\bar{\theta}}(s',a')
\nonumber -\alpha\log \pi(a'|s'))))^{2}]
\end{align}
}
where $\alpha > 0$ is the trade-off coefficient of the entropy regularization term and $\gamma$ is the discount factor. 
$Q_{\bar{\theta}}$ is the target network~\cite{haarnojaSoftActorCriticOffPolicy2018}.
However, this objective is written in terms of combined policy~$\pi$ and the environment action $a$. We want to change this objective to use the latent action $\hat{a}$ instead of $a$, so we can avoid the need to backpropagate through the mapping function during RL training.

\begin{prop}
The log-probability of the combined policy, \(\log \pi(a|s)\), can be approximated using $\hat{a}$ as:
\[
\log \pi(a|s) \approx \log \mu_\phi(\hat{a}|s) + \frac{\|\hat{a}\|_2^2}{2} + K(s) 
\]
\end{prop}
\noindent where $K(s)$ is a constant independent of the action $a$.

\begin{proof}
We first apply the change of variables theorem to the combined policy, which results in:
\begin{equation}\label{eq:change_of_v_pi}
\log \pi(a|s) = \log \mu_\phi(\hat{a}|s) - \log |\det J_{f_\psi}(\hat{a};s)|    
\end{equation}

% This still involves the Jacobian of the mapping function. 
% Then we eliminate the Jacobian term by considering another property of the trained mapping function: the trained model \(q(a|s)\) Eq.~\eqref{eq:conditional_nvp_prob} should approximate the target distribution \(p(a|s)\)
Then we eliminate the Jacobian term by considering another property of the trained mapping function: the trained model \(q(a|s)\) should approximate the target distribution \(p(a|s)\)
\red{which is a uniform distribution over feasible environment actions as noted in Eq.~\eqref{eq:target_distribution}} (i.e. $\log p(a|s) \approx \log q(a|s)$).
% Then if we apply change of variable theorem on \textcolor{red}{it: refer to what} we get,
By substituting  $\log q(a|s)$ from Eq.~\eqref{eq:conditional_nvp_prob} we get,
\begin{equation}\label{eq:change_of_v_p}
\log p(a|s) \approx \log \hat{q}(\hat{a}) - \log |\det J_{f_\psi}(\hat{a};s)|    
\end{equation}

Then we subtract Eq.~\eqref{eq:change_of_v_p} from Eq.~\eqref{eq:change_of_v_pi} to eliminate the common term $\log |\det J_{f_\psi}(\hat{a};s)| $, which results in:
\begin{equation}\label{eq:log_pi}
\log \pi(a|s) \approx \log \mu_\phi(\hat{a}|s) + \log p(a|s) - \log \hat{q}(\hat{a})    
\end{equation}

Here, the base distribution $\hat{q}(\hat{a})$ is a standard Gaussian ($d$-dimensional) and $p(a|s)$ is constant for all feasible actions given $s$. Thus, we have, 
% \vskip -1pt
{
\small
\begin{align}\label{eq:log_pi_final}
\log \pi(a|s) &\approx \log \mu_\phi(\hat{a}|s) +\log p(a|s) -\log \frac{1}{\sqrt{(2\pi)^d}} e^{ -\frac{\hat{a}^T \hat{a}}{2} }\nonumber 
\\
&\approx \log \mu_\phi(\hat{a}|s) +\log p(a|s)  + \log \sqrt{(2\pi)^d} + \frac{\|\hat{a}\|_2^2}{2} \nonumber
\\
& \approx \log \mu_\phi(\hat{a}|s) + \frac{\|\hat{a}\|_2^2}{2} + K(s)  
\end{align}
}
\end{proof}

\noindent \textbf{Loss Functions:}
We substitute $\log\pi(a|s)$ using Eq.~\eqref{eq:log_pi_final} in the objective Eq.~\eqref{eq:sac_policy_loss_original} and Eq.~\eqref{eq:sac_critic_loss_original} to get our final objectives for policy and critic update. 
 \( K(s) \) can be ignored as it is constant for all feasible actions.

% \vskip -1pt
{
\small
\begin{equation}
J^{\mu}(\phi)=-\!\!\!\!\!\!\!\!\!\!\!\!\E_{s\sim \mathcal{B}, \hat{a}\sim \mu_\phi(\cdot|s)}
\!\!\!\!\!\!\!\!\!\!\!\![Q^{\mu_\phi}(s,\hat{a})-\alpha(\log \mu_\phi(\hat{a}|s)+\frac{\|\hat{a}\|^2_2}{2})]\label{eq:sac_policy_loss}
\end{equation}
}
% \vskip -1pt
{
\small
\begin{align}\label{eq:sac_critic_loss}
J^{Q}(\theta)=\!&\E_{\substack{(s,\hat{a},r,s')\sim \mc{B}, \hat{a}'\sim \mu_\phi(\cdot|s')}}[(Q_{\theta}(s,\hat{a}) -(r+\gamma (Q_{\bar{\theta}}(s',\hat{a}')
\nonumber \\
& \qquad \qquad -\alpha\log (\mu_\phi(\hat{a}'|s')+\frac{\|\hat{a}'\|^2_2}{2} ))
))^{2}]
\end{align}
}
\noindent For notation simplicity, we ignore the reparameterization~\cite{kingmaAutoEncodingVariationalBayes2022} for sampling actions in the objective. 
However, we do use this when optimizing the parameters $\phi$.

\section{Experimental Results}

\begin{figure*}[tb]
  \centering	    
  \includegraphics[width=0.75\textwidth]{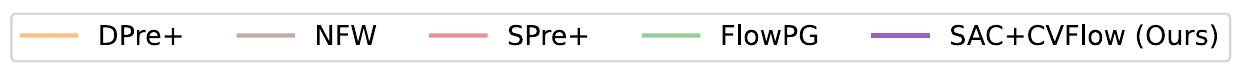}
  % \vspace{-0.2cm}
  \includegraphics[width=\linewidth]{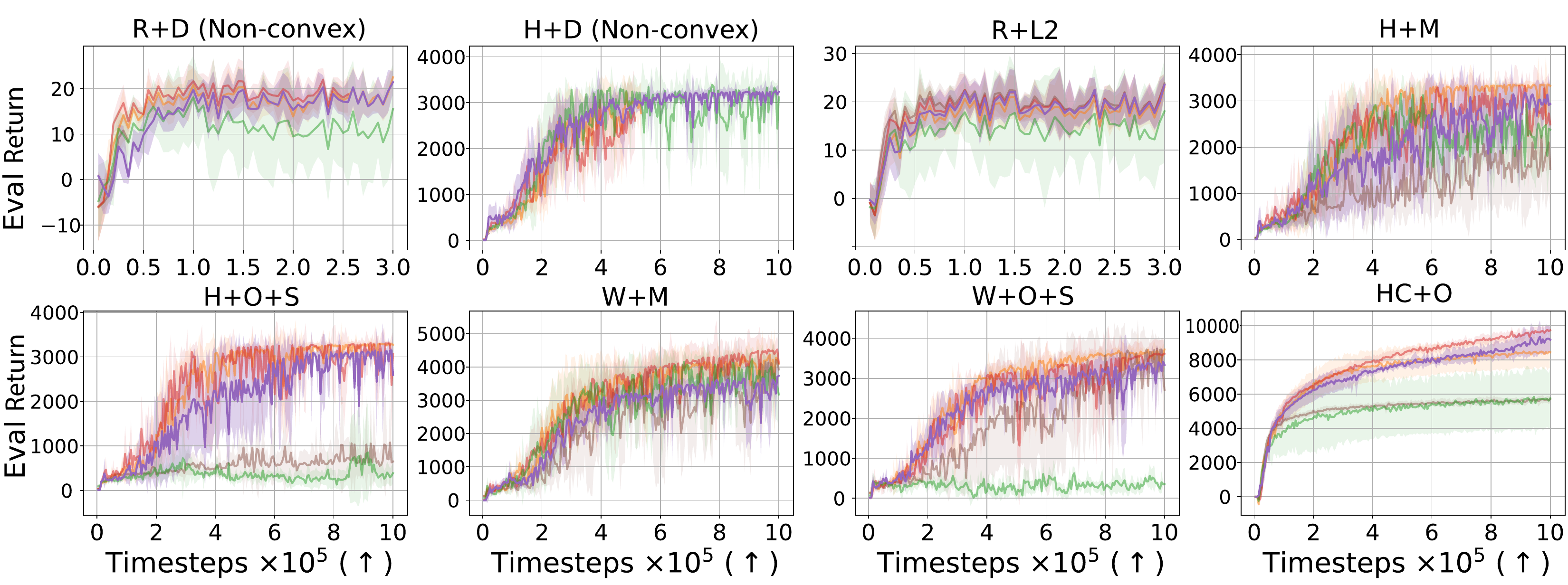}
  \caption{Evaluation returns for eight MuJoCo continuous control tasks during training. A higher return is better.}
  \label{fig:rl_results}
  % \vspace{-0.3cm}
\end{figure*}

The goal of our experiments is to: 
(1) evaluate whether our approach results in fewer constraint violations during training compared to other approaches, without sacrificing returns; 
(2) whether {\our} can be adapted successfully for state-wise constraints where the analytical form of the action constraints is not available;
(3) \red{assess whether the \blue{$\|a\|^2_2$} term in the entropy regularization segment of SAC is truly beneficial};
and 
(4) determine whether {\our} yields higher accuracy while covering most of the feasible region compared to the standard method of training the flow using a sampled dataset of feasible environment actions; 

\noindent \textbf{Action-constrained environments:} 
We evaluate our approach on four MuJoCo~\cite{todorovMuJoCoPhysicsEngine2012} continuous control environments: Reacher (R), Hopper (H), Walker2D (W), and HalfCheetah (HC). Using action constraints from previous work~\cite{Kasaura23}, we establish eight constrained control tasks: \emph{R+L2}, \emph{H+M}, \emph{H+O+S}, \emph{W+M}, \emph{W+O+S}, \emph{HC+O}, \emph{R+D}, and \emph{H+D}. These constraints restrict joint movement in each task, with details in Table 1 of the supplementary. \emph{R+D} and \emph{H+D} are non-convex constraints derived from modified convex constraints in~\cite{Kasaura23}, aimed at evaluating the efficiency of projection-based methods for challenging non-convex problems.

\noindent \textbf{State-constrained environments:} We evaluate our approach on four continuous control tasks with state-wise constraints: \emph{Ball1D}, \emph{Ball3D}, \emph{Space-Corridor}, and \emph{Space-Arena}, as proposed in previous work~\cite{dalal2018safe}. In \emph{Ball1D} and \emph{Ball3D}, the goal is to move a ball as close as possible to a target by adjusting its velocity within safe regions of $[0,1]$ and $[0,1]^3$, respectively. In \emph{Space-Corridor} and \emph{Space-Arena}, the task is to navigate a spaceship to a target location by controlling thrust engines within a two-dimensional universe, avoiding walls. Unlike in action constrained environments, episodes terminate upon violation of state-wise constraints.

% \vskip -1pt
{
\begin{table}[tb]
        \begin{center}
                \begin{tabular}{|l|c|c|c|c|c|}
                        \hline
                        Problem & DPre+ & NFW & SPre+ & FlowPG & Ours \\
                        \hline
                        R+D & 98.15 & 95.50 & 97.03 & 24.79 & \textbf{0.01} \\
                        \hline
                        H+D & 74.10 & 74.89 & 77.72 & 32.29 & \textbf{2.18} \\
                        \hline
                        R+L2 & 82.51 & 22.71 & 16.47 & \textbf{0.03} & 1.70 \\
                        \hline
                        H+M & 3.67 & 4.45 & 3.25 & 4.93 & \textbf{0.25} \\
                        \hline
                        H+O+S & 42.44 & \textbf{2.14} & 7.83 & 53.91 & {2.42} \\
                        \hline
                        W+M & 30.55 & 4.50 & 11.70 & 16.40 & \textbf{2.41} \\
                        \hline
                        W+O+S & 84.89 & 3.00 & 20.44 & 47.80 & \textbf{1.55} \\
                        \hline
                        HC+O & 73.57 & 9.73 & 46.66 & 61.10 & \textbf{5.04} \\
                        \hline\hline
                        Ball1D & \textbf{0.00} & \textbf{0.00} & \textbf{0.00} & \textbf{0.00} & \textbf{0.00} \\
                        \hline  
                        Ball3D & 16.01 & \textbf{0.00} & 23.16 & 4.07 & 0.37 \\
                        \hline   
                        SpaceC & 54.32 & 89.54 & 23.71 & 59.07 & \textbf{12.06} \\
                        \hline  
                        SpaceA & 51.65 & 91.74 & 13.60 & 29.32 & \textbf{10.78} \\
                        \hline
                \end{tabular}
        \end{center}
        % \vspace{-0.4cm}
        \caption{
        %The percentage of training steps during which the agent produced infeasible actions, requiring the use of an analytical solver to project the action.
        The percentage of constraint violations during RL training. A lower value is preferable. The standard deviation of the constraint violations is reported in the supplementary. The abbreviations SpaceC and SpaceA refer to the Space-Corridor and Space-Arena benchmarks, respectively.
        }
        % \vspace{-0.3cm}
        \label{tab:cv-count}
\end{table}
}

\begin{figure*}[tb]
    \centering	
    \includegraphics[width=0.75\linewidth]{figures/analytical-legend.pdf}\\
    \includegraphics[width=\linewidth]{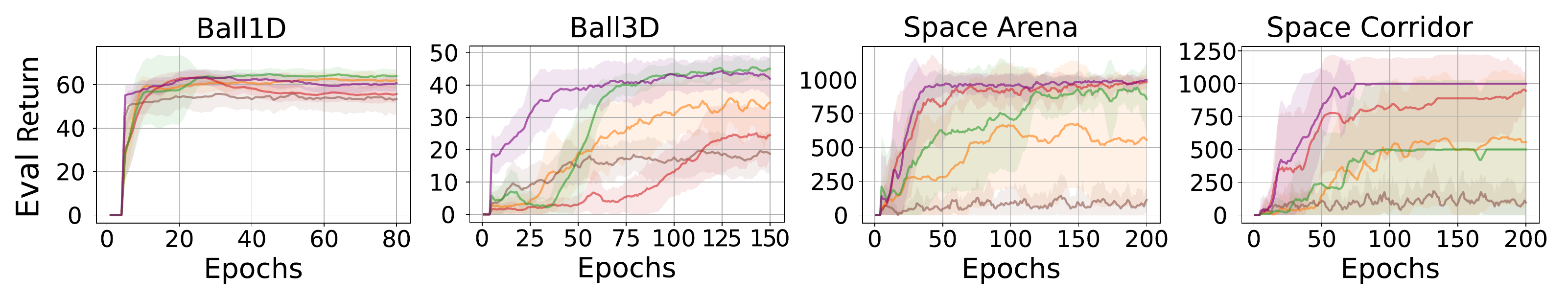}
    \caption{Evaluation returns for four state-constrained tasks during training. A higher return ($\uparrow$) is better.} 
    \label{fig:state-wise-rewards}
% \vspace{-0.2cm}
\end{figure*}

\begin{figure}[tb]
    \centering	
    \includegraphics[width=\linewidth]{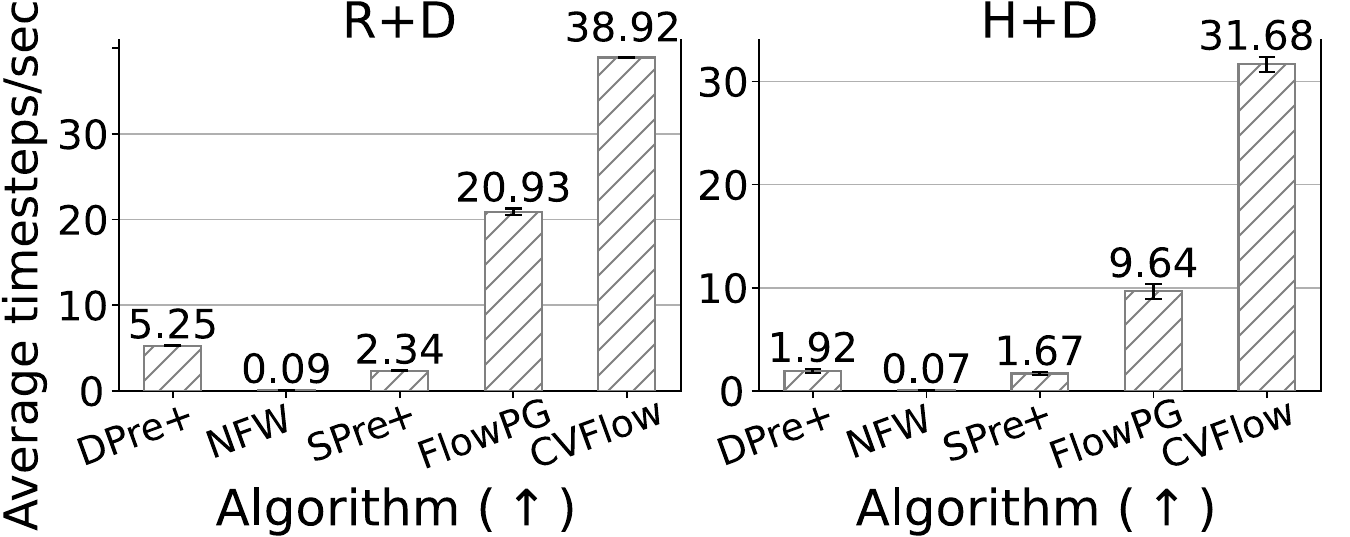}
    \caption{Average timesteps per second of the RL agent for non-convex constraints tasks (higher is better $\uparrow$), CVFlow based approach has a significantly higher frame rate.} 
    \label{fig:timesteps-non-convex}
    % \vspace{-.5cm}
\end{figure}

\noindent \textbf{Baselines:} We integrated \our\ with a standard Gaussian base distribution with SAC as described in the Section~\ref{sec:sac_cv_flow}, referring to it as \textbf{SAC+CVFlow}. Our approach is compared with four baseline algorithms, \textbf{DPre+}, \textbf{SPre+}, \textbf{NFW} and \textbf{FlowPG} which have shown best results in previous studies \cite{Kasaura23,lin2021escaping, brahmanage2023flowpg}. DPre+ is a DDPG-based algorithm with an additional projection step, ensuring that the projected action is feasible. SPre+ is same to DPre+ except for the underlying RL algorithm, which is SAC. Both DPre+ and SPre+ employ a penalty term based on the constraint violation signal. Neural Frank Wolfe (NFW) Policy Optimization~\cite{lin2021escaping} is also a DDPG-based algorithm. It leverages the Frank-Wolfe algorithm to update the policy network instead of the standard policy gradient. FlowPG~\cite{brahmanage2023flowpg} is built upon DDPG and uses normalizing flows to map the latent action into the feasible region. However, unlike our method, FlowPG requires feasible state-action pairs for training the normalizing flows and does not use the maximum entropy objective with RL.
\blue{All algorithms include a projection step to correct actions if they are infeasible, as ACRL requires only feasible actions. In state-constrained environments, where action constraints are linearly approximated (as discussed in Section~\ref{sec:state-wise}), the projection step might still produce infeasible actions, which can result in the termination of the episode.
}

Each algorithm is trained with 10 random seeds, capped at 48 hours per run, using hyperparameters and architectures from \cite{Kasaura23} (details in supplementary material).

\subsection{Performance on MuJoCo Tasks}

\noindent \textbf{Reward comparisons: }
Evaluation returns are computed by running five episodes per random seed every 5k training steps. Figure~\ref{fig:rl_results} shows that our approach SAC+CVFlow achieves comparable results to DPre+ and SPre+ in terms of evaluation return across all the tasks. It finds high-quality policies that yield good returns with fewer constraint violations as discussed next. 

\noindent \textbf{Constraint violations: }
\blue{We measured the percentage of timesteps during the training period in which the agent produced infeasible actions before the projection step.}
This is equal to the number of QP-solver calls because each infeasible action must be projected into the feasible region before it is executed in the environment. As shown in Table~\ref{tab:cv-count} our approach results in fewer constraint violations across all tasks compared to all the baselines, except for (H+O+S, R+L2). In some cases (i.e., H+M, R+D, H+D), our approach even achieves a reduction in constraint violations by an order-of-magnitude compared to all the baselines. In the case of R+L2, we also observe an order-of-magnitude reduction in constraint violations compared to DPre+, SPre+, and NFW. Even in cases where our approach ranks second (H+O+S, R+L2), it yields comparable results to that of the best baseline. Furthermore, we observe that our method not only has fewer constraint violations but the magnitude of these violations is also lower as shown in the Figure~\ref{sup:cv-magnitude} of the supplementary. This indicates that even when the constraints are violated the produced infeasible action is still closer to the feasible region. 

\noindent \textbf{Runtime: }In terms of timesteps per second, our algorithm generally achieves comparable results to baseline methods, except for the non-convex tasks R+D and H+D, where it demonstrates significantly faster runtime, as shown in Figure~\ref{fig:timesteps-non-convex}. For non-convex tasks, our approach achieves an order of magnitude faster runtime compared to DPre+, SPre+ and NFW, while achieving at least twice as fast as FlowPG. The overhead cost in DPre+, SPre+, and NFW arises from using solvers to find feasible actions that satisfy non-convex constraints. FlowPG, on the other hand, incurs significant runtime due to backpropagation through the normalizing flow during training. 
Results for timesteps per second on other tasks can be found in Figure~\ref{sup:timesteps} of the supplementary material, along with computing infrastructure details.

\subsection{Performance on State-Wise Constraints}
\noindent \textbf{Reward comparisons: }
\blue{
The advantage of our approach is further demonstrated in state-constrained environments. As shown in Figure~\ref{fig:state-wise-rewards}, our approach outperforms baseline methods in terms of return across all environments, except for Ball1D, where it still achieves comparable results.
}

\noindent \textbf{Constraint violations: }
\blue{
Since the analytical form of action constraints is not available for these tasks, it is not accurate to measure pre-projection constraint violations. 
Therefore, we report the percentage of post-projection constraint violations, which corresponds to the percentage of episode terminations due to state-wise constraint violations. Table~\ref{tab:cv-count} presents the constraint violation results. In Ball1D, all algorithms show no constraint violations, as the constraints are relatively easy to handle compared to other tasks. Our approach results in fewer constraint violations across all tasks, except for Ball3D. Although NFW has fewer violations in Ball3D, it only converged to a significantly lower return.
}
\subsection{The Effect of the Entropy Term}

In Figure~\ref{fig:ablation_penalty} of the supplementary, we evaluate whether the proposed entropy term in Eq.~\eqref{eq:sac_critic_loss} and Eq.~\eqref{eq:sac_policy_loss} has a meaningful effect on SAC+CVFlow algorithm with a Gaussian base distribution. Therefore, we show the return and constraint violation percentage of the algorithm, with and without the entropy term ($\|\hat{a}\|_2^2$). We can see that without the entropy term, the agent produces higher constraint violations (except for H+M). Additionally, without the entropy term, the agent struggles to learn in all environments.

\subsection{Comparison of Flow Models}\label{sec:comp_flow}

\blue{
We also compare the quality of the trained normalizing flows using our proposed \ourcv-Flow with the method from~\cite{brahmanage2023flowpg}, which relies on feasible state-action pairs. As shown in supplementary Figure~\ref{fig:compare-flow}, we evaluated accuracy, recall, and F1-score during training. The key observation is that \ourcv-Flow achieves higher accuracy and a better F1-score compared to normalizing flows trained on a feasible dataset. This higher accuracy allows our flow model to output feasible actions with high probability, reducing the need for QP-based action projections, which are time-consuming when integrated with ACRL methods.
}

%%%%%%%%%%%%%%%%%%%%%%%%%%%%%%%%%%%%%%%%%%%%%%%%%%%%%%%%%%%%%%%%%%%%%%%%

%%% The acknowledgments section is defined using the "acks" environment
%%% (rather than an unnumbered section). The use of this environment 
%%% ensures the proper identification of the section in the article 
%%% metadata as well as the consistent spelling of the heading.

\section{Conclusion}
We have introduced a normalizing flows-based mapping function that transforms samples from a base distribution into feasible actions in ACRL. A key advantage of our flow model is that it eliminates the need to generate feasible state-action samples from the constrained action space for training, which is often challenging. Instead, our model is trained using constraint violation signals. When integrated with SAC to address both action and state-wise constraints, our approach results in significantly fewer constraint violations without sacrificing returns, compared to previous methods. Additionally, it incurs less overhead when handling non-convex constraints.

\newpage
\section*{Acknowledgments}
This research/project is supported by the National Research Foundation, Singapore and DSO National Laboratories under the AI Singapore Programme (AISG Award No: AISG2- RP-2020-017).

% \section{Acknowledgments}
% AAAI is especially grateful to Peter Patel Schneider for his work in implementing the original aaai.sty file, liberally using the ideas of other style hackers, including Barbara Beeton. We also acknowledge with thanks the work of George Ferguson for his guide to using the style and BibTeX files --- which has been incorporated into this document --- and Hans Guesgen, who provided several timely modifications, as well as the many others who have, from time to time, sent in suggestions on improvements to the AAAI style. We are especially grateful to Francisco Cruz, Marc Pujol-Gonzalez, and Mico Loretan for the improvements to the Bib\TeX{} and \LaTeX{} files made in 2020.

% The preparation of the \LaTeX{} and Bib\TeX{} files that implement these instructions was supported by Schlumberger Palo Alto Research, AT\&T Bell Laboratories, Morgan Kaufmann Publishers, The Live Oak Press, LLC, and AAAI Press. Bibliography style changes were added by Sunil Issar. \verb+\+pubnote was added by J. Scott Penberthy. George Ferguson added support for printing the AAAI copyright slug. Additional changes to aaai25.sty and aaai25.bst have been made by Francisco Cruz, Marc Pujol-Gonzalez, and Mico Loretan.

% \bigskip
% \noindent Thank you for reading these instructions carefully. We look forward to receiving your electronic files!

\bibliography{aaai25}
\appendix
\onecolumn

\newpage
\section{Algorithm for Training SAC with CV Flow}\label{sup:fine_tune}
\begin{algorithm}[ht]
\caption{Soft Actor Critic with \our}
\label{alg:soft_actor_critic}
\begin{algorithmic}[1]
\STATE Load pre-trained flow model: $f_\psi$
\STATE Initialize critics and policy parameters: $\theta$, $\bar{\theta}$,$\phi$
\STATE $\mc{B} \la \varnothing$ \COMMENT{Initialize empty replay buffer}
\FOR{each iteration}
	\FOR{each environment step}
	\STATE $\hat{a}_t \sim \mu_\phi(\hat{a}_t|s_t)$ \COMMENT{Sample latent action}
	\STATE $\hat{a}_t \leftarrow  \min(\hat{a}_t, min=-3, max=3)$ \COMMENT{Clip the latent action at 3-sigma}
        \STATE $a_t \la f_\psi(\hat{a}_t,s_t)$ \COMMENT{Apply flow to get environment action}
        \IF{$a_t \notin \mc{C}(s_t)$ }
            \STATE $a_t \la \argmin_{\tilde{a}_t \in \mc{C}(s_t)} \|\tilde{a}_t - a_t\|^2_2$ 
        \ENDIF
	\STATE $s_{t+1} \sim p(s_{t+1}| s_t, a_t)$ \COMMENT{Collect next state}
	\STATE $\mathcal{B} \la \mathcal{B} \cup \left\{\langle s_t, \hat{a}_t, r(s_t, a_t), s_{t+1}\rangle\right\}$
	\ENDFOR
 
	\FOR{each gradient step}
        \STATE $\theta \la \theta - \lambda_Q \hat\nabla_\theta J^Q(\theta)$ \COMMENT{Update Q function using Eq.\eqref{eq:sac_critic_loss}}
        \STATE $\phi \la \phi - \lambda_\mu \hat\nabla_\phi J^\mu(\phi)$ \COMMENT{Update policy using Eq.\eqref{eq:sac_policy_loss}}
        % \STATE $\alpha \la \alpha - \lambda_\alpha \hat\nabla J(\alpha)$ \COMMENT{Update temperature}
        \STATE $\bar{\theta} \la \tau \theta + (1-\tau){\bar{\theta}} $ \COMMENT{Update target network}
	\ENDFOR
\ENDFOR
\end{algorithmic}
\end{algorithm}

\section{Algorithm for Training DDPG with CV Flow}\label{sup:ddpg}
Our \ourcv-Flow can also be integrated with the DDPG algorithm~\cite{lillicrap2015continuous}, where the policy is implemented as a deterministic function of the state. The pseudocode for the algorithm is provided in Algorithm~\ref{alg:ddpg}.
\vskip -1pt
{
\small
\begin{equation}
\nabla_\phi J^{\mu}(\phi) = \E_{s\sim \mathcal{B}} \nabla_a Q(s, a)  \nabla_{\hat{a}}  f_\psi(\hat{a},s) \nabla_\phi \mu_\phi(s)|_{\hat{a}=\mu_\phi(s),a=f_\psi(\hat{a},s)}    \\
\label{eq:ddpg_policy_loss}
\end{equation}
}
\vskip -1pt
{
\small
\begin{align}\label{eq:ddpg_critic_loss}
J^{Q}(\theta)=\E_{\substack{(s,{a},r,s')\sim \mc{B}}}
[(
Q_{\theta}(s,{a}) -(r+\gamma Q_{\bar{\theta}}(s',a'
)
)
)^{2}|_{a' = f_\psi(\mu_{\phi}(s'), s')}]
\end{align}
}
\begin{algorithm}[ht]
\caption{ DDPG with \our}
\label{alg:ddpg}
\begin{algorithmic}[1]
\STATE Load pre-trained flow model: $f_\psi$
\STATE Initialize critics and policy parameters: $\theta$, $\bar{\theta}$,$\phi$
\STATE $\mc{B} \la \varnothing$ \COMMENT{Initialize empty replay buffer}
\FOR{each iteration}
	\FOR{each environment step}
	\STATE $\hat{a}_t \leftarrow \mu_\phi(s_t)$ \COMMENT{Get latent action from the deterministic policy}
        \STATE $\hat{a}_t \leftarrow \hat{a}_t + \delta$ \COMMENT{Add noise \(\delta \sim \mathcal{N}(0,1)\) to the action for exploration.} 
	\STATE $\hat{a}_t \leftarrow  \min(\hat{a}_t, min=-3, max=3)$ \COMMENT{Clip the latent action to stay within 3-sigma}
        \STATE $a_t \la f_\psi(\hat{a}_t,s_t)$ \COMMENT{Apply flow to get environment action}
        \IF{$a_t \notin \mc{C}(s_t)$ }
            \STATE $a_t \la \argmin_{\tilde{a}_t \in \mc{C}(s_t)} \|\tilde{a}_t - a_t\|^2_2$ 
        \ENDIF
	\STATE $s_{t+1} \sim p(s_{t+1}| s_t, a_t)$ \COMMENT{Collect next state}
	\STATE $\mathcal{B} \la \mathcal{B} \cup \left\{\langle s_t, {a}_t, r(s_t, a_t), s_{t+1}\rangle\right\}$
	\ENDFOR
 
	\FOR{each gradient step}
        \STATE $\theta \la \theta - \lambda_Q \nabla_\theta J^Q(\theta)$ \COMMENT{Update Q function using Eq.~\eqref{eq:ddpg_critic_loss}}
        \STATE $\phi \la \phi - \lambda_\mu \nabla_\phi J^\mu(\phi)$ \COMMENT{Update policy using Eq.~\eqref{eq:ddpg_policy_loss}}
        % \STATE $\alpha \la \alpha - \lambda_\alpha \hat\nabla J(\alpha)$ \COMMENT{Update temperature}
        \STATE $\bar{\theta} \la \tau \theta + (1-\tau){\bar{\theta}} $ \COMMENT{Update target network}
	\ENDFOR
\ENDFOR
\end{algorithmic}
\end{algorithm}

\section{Experiment Setup}

Here we present the exact constraints of our tasks in Table~\ref{sup:envs}. To train the flow, the state distribution $p_S$ was determined based on the environment. Specifically, to train the flow, we do not need the complete state; only the component of the state distribution that the constraint depends on is required. We observed that most of these state variables are bounded, except for HC+O. Therefore, the state distribution was defined as a uniform distribution over the bounded space. The standard deviation of $w_i$ for HC+O was calculated based on the data collected by running an unconstrained agent on the environment.

{
\begin{table*}[htb]
\begin{center}
\setlength\extrarowheight{3.5pt}
\begin{tabular}{l c l c r}
 Environment & Name & Constraint & Convexity & State Distribution ($p_S$) \\ 
 \hline 
Reacher & R+L2 & $a_1^2 + a_2^2 \leq 0.05$ & Convex & N/A \\  
 & R+D & $0.04 \leq a_1^2 + a_2^2 \leq 0.05$ & Non-Convex & N/A\\
 % \hline
 Hopper & H+M & $ \sum_{i=1}^3 max\{w_i a_i, 0\}  \leq 10$ & Convex & $w_i \sim \mc{U}_{[-10,10]}$ \\
  & H+O+S & $\sum_{i=1}^3 |w_i a_i|  \leq 10 \land \sum_{i=1}^3 a_i^2 sin^2 \theta_i  \leq 0.1$ & Convex 
  & $w_i \sim \mc{U}_{[-10,10]}, \theta_i \sim \mc{U}_{[-\pi,\pi]}$
  \\
  & H+D & $1.4 \leq \sum_{i=1}^3a_i^2 \leq 1.5$ & Non-Convex & N/A \\
  % \hline
Walker2D & W+M & $ \sum_{i=1}^6 max\{w_i a_i, 0\}  \leq 10$ & Convex & $w_i \sim \mc{U}_{[-10,10]}$ \\
  & W+O+S & $\sum_{i=1}^6 |w_i a_i|  \leq 10 \land \sum_{i=1}^6 a_i^2 sin^2 \theta_i  \leq 0.1$ & Convex
  & $w_i \sim \mc{U}_{[-10,10]}, \theta_i \sim \mc{U}_{[-\pi,\pi]}$ \\
HalfCheetah & HC+O & $\sum_{i=1}^6 |w_i a_i|  \leq 20$ & Convex & $w_i \sim \mc{N}_{\mu=0, \sigma=15}$ \\
\end{tabular}
\end{center}
\caption{Analytical expressions of constraints and their environments.}\label{sup:envs}
\end{table*}

\begin{table*}[ht]
\begin{center}
\setlength\extrarowheight{2.5pt}
\begin{tabular}{l c c c c c c c c}
Hyperparameters & \multicolumn{2}{c}{Reacher}  &  \multicolumn{2}{c}{Hopper} &  \multicolumn{2}{c}{Walker2D/HalfCheetah} & BallND/Spaceship \\
\hline
& DDPG & SAC & DDPG & SAC & DDPG & SAC & DDPG  \\
Discount factor & 0.98 & 0.98 & 0.99 & 0.99 & 0.99 & 0.99 & 0.99 \\
Net. arch. 1st layer & 400 &  400 &  400 & 256 & 400 & 256 & 64  \\
Net. arch. 2nd layer & 300 &  300 &  300 & 256 & 300 & 256 & 64  \\
Batch size & 100 & 256 & 256 & 256 & 100 &  256 & 256  \\
Learning starts & 1e5 & 1e5 & 1e5 & 1e5 & 1e5 & 1e5 & 256  \\
FW Learning rate & 0.05 & - & 0.01 & - & 0.01 & - & 0.05  \\
Learning rate & 1e-3 & 7.3e-4 & 3e-4 & 3e-4 & 1e-3 & 3e-4 & 1e-4  \\
\end{tabular}
\end{center}
\caption{Hyperparameters: SPre+ and \our, which are based on SAC, utilize the hyperparameters listed in the SAC column. Conversely, DPre+, NFW and FlowPG based on DDPG, uses the hyperparameters specified in the DDPG column.}
\end{table*}
}

\section{Results}

\begin{table}
        \begin{center}
                \begin{tabular}{|l|l|l|l|l|l|}
                        \hline
                        Problem & DPre+ & NFW & SPre+ & FlowPG & SAC+CVFlow (Ours) \\
                        \hline
                        R+D (Non-convex) & 98.15 ± 0.18 & 95.50 ± 0.09 & 97.03 ± 0.04 & 24.79 ± 10.27 & 0.01 ± 0.00 \\
                        \hline
                        H+D (Non-convex) & 74.10 ± 13.27 & 74.89 ± 9.62 & 77.72 ± 13.77 & 32.29 ± 3.58 & 2.18 ± 1.05 \\
                        \hline
                        R+L2 & 82.51 ± 3.48 & 22.71 ± 0.30 & 16.47 ± 0.37 & 0.03 ± 0.00 & 1.70 ± 0.22 \\
                        \hline
                        H+M & 3.67 ± 2.68 & 4.45 ± 1.56 & 3.25 ± 1.38 & 4.93 ± 1.07 & 0.25 ± 0.20 \\
                        \hline
                        H+O+S & 42.44 ± 6.35 & 2.14 ± 0.94 & 7.83 ± 1.16 & 53.91 ± 6.26 & 2.42 ± 1.19 \\
                        \hline
                        W+M & 30.55 ± 14.89 & 4.50 ± 1.78 & 11.70 ± 4.14 & 16.40 ± 5.95 & 2.41 ± 0.72 \\
                        \hline
                        W+O+S & 84.89 ± 7.48 & 3.00 ± 0.62 & 20.44 ± 3.70 & 47.80 ± 19.10 & 1.55 ± 1.20 \\
                        \hline
                        HC+O & 73.57 ± 4.82 & 9.73 ± 2.40 & 46.66 ± 1.68 & 61.10 ± 5.21 & 5.04 ± 1.21 \\
                        \hline\hline
                        Ball1D & 0.00 ± 0.00 & 0.00 ± 0.00 & 0.00 ± 0.00 & 0.00 ± 0.00 & 0.00 ± 0.00 \\
                        \hline  
                        Ball3D & 16.01 ± 11.82 & 0.00 ± 0.00  & 23.16 ± 17.72 & 4.07 ± 4.16 & 0.37 ± 0.68 \\
                        \hline   
                        Space Corridor & 54.32 ± 31.63 & 89.54 ± 2.48 & 23.71 ± 26.31 & 59.07 ± 38.09 & 12.06 ± 6.30 \\
                        \hline  
                        Space Arena & 51.65 ± 29.59 & 91.74 ± 1.95 & 13.60 ± 6.01 & 29.32 ± 17.41 & 10.78 ± 4.07 \\
                        \hline
                \end{tabular}
        \end{center}
        \caption{The percentage of constraint violations during RL training. A lower value is preferable.}
        \label{tab:--}
\end{table}

\begin{table}[tb]
\begin{center}
\begin{tabular}{|l|l|l|l|l|l|}
\hline
Problem          & DPre+         & NFW           & SPre+         & FlowPG        & SAC+CVFlow (Ours) \\ \hline
R+D (Non-convex) & 65.70 ± 0.16  & 94.92 ± 0.38  & 96.74 ± 0.48  & 24.89 ± 10.08 & 0.00 ± 0.00       \\ \hline
H+D (Non-convex) & 85.00 ± 18.76 & 69.47 ± 24.94 & 75.27 ± 21.87 & 27.81 ± 8.83  & 2.62 ± 1.86       \\ \hline
R+L2             & 55.40 ± 2.87  & 22.22 ± 1.74  & 13.56 ± 1.57  & 0.00 ± 0.00   & 1.40 ± 0.51       \\ \hline
H+M              & 5.37 ± 3.31   & 5.82 ± 3.81   & 4.70 ± 2.21   & 8.26 ± 2.66   & 0.36 ± 0.24       \\ \hline
H+O+S            & 58.46 ± 4.83  & 2.45 ± 1.49   & 7.77 ± 1.08   & 35.52 ± 16.81 & 4.63 ± 1.73       \\ \hline
W+M              & 55.10 ± 22.17 & 4.08 ± 2.03   & 10.86 ± 6.72  & 32.62 ± 13.35 & 1.75 ± 0.80       \\ \hline
W+O+S            & 90.94 ± 17.44 & 2.19 ± 1.34   & 18.04 ± 4.20  & 63.77 ± 40.43 & 0.85 ± 0.83       \\ \hline
HC+O             & 71.92 ± 5.54  & 10.33 ± 2.40  & 43.22 ± 4.95  & 60.59 ± 8.35  & 5.32 ± 1.73       \\ \hline   
\end{tabular}
        \caption{Percentage of constraint violations, averaged over 100 episodes, when executing the final trained policy.}
        \label{tab:cv-at-runtime}
\end{center}
\end{table}

\begin{figure*}[tb]
    \centering	
    \includegraphics[width=0.3\linewidth]{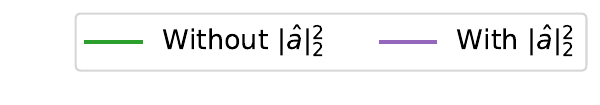}\\
    \vspace{-0.2cm}
    \includegraphics[width=0.75\linewidth]{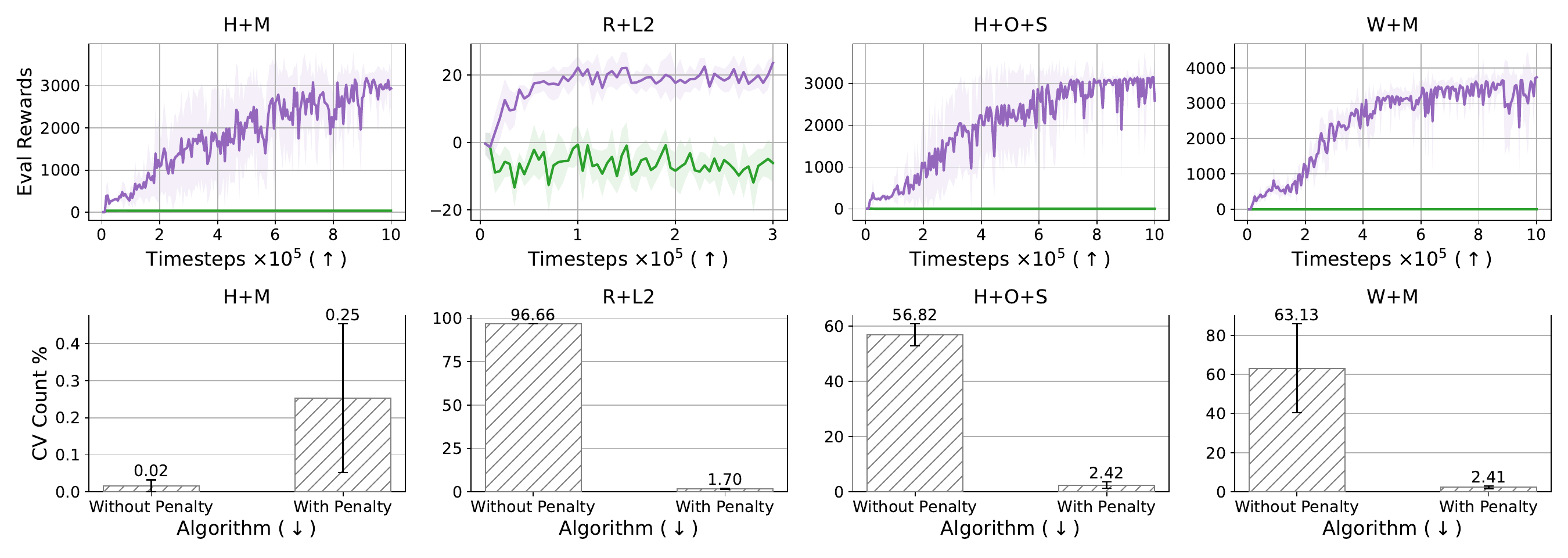}
    \vspace{-0.2cm}
    \caption{Comparison of results for the flow with and without the $\|\hat{a}\|_2^2$ term for RL agent is shown in this figure. The first row contains the return of the agent, while the second row shows the constraint violation (CV) count as a percentage of total-timesteps. Without the $\|\hat{a}\|_2^2$  term, the agent not only produce higher constraint violations but also struggles to learn.}\label{fig:ablation_penalty}
    \vspace{-0.4cm}
\end{figure*}

\subsection{Comparison of Standard Flow and \our}
Here, we compare our approach against standard flow in terms of  the accuracy, recall and F1-score during the training process. 

\noindent\textbf{Accuracy} is the percentage of samples generated using normalizing flow that fall within the feasible region. To evaluate the accuracy of the model, we first generate $n$ samples using the flow and then calculate the percentage of these samples that satisfy predefined action constraints. 

\noindent\textbf{Recall} (also called coverage) indicates the fraction of valid actions that can be generated from the latent space. To measure the recall we first generate samples from the feasible region using a technique such as rejection sampling. Then we map these samples to the latent space using the inverse of the flow model $f^{-1}$ and compute the percentage that falls within the domain of the latent space $[-1, 1]^d$. Mathematically, given a conditioning variable $s$ and constrained space $\mathcal{C}(s)$, the recall is computed as follows. 
{
\begin{equation} \label{eq:recall}
recall(s) = \frac{\sum_{a \in \mathcal{C}(s)} \mathbb{I}_{dom_f}f^{-1}(a|s)}{|\mathcal{C}(s)|}
\end{equation}}

\noindent\textbf{F1-Score} is the standard machine
learning evaluation metric, which combines accuracy and recall.

\noindent As presented in Figure~\ref{fig:compare-flow} \ourcv-Flow achieves higher accuracy and a better F1 score compared to the standard flow. We use a uniform base distribution for this experiment because a Gaussian base distribution, with its support $[-\infty, \infty]$, does not allow exact measurement of coverage.

\begin{figure}[b]
  \centering	
  \includegraphics[width=0.5\linewidth]{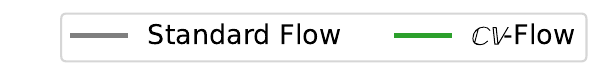}\\
  \includegraphics[width=0.99\linewidth]{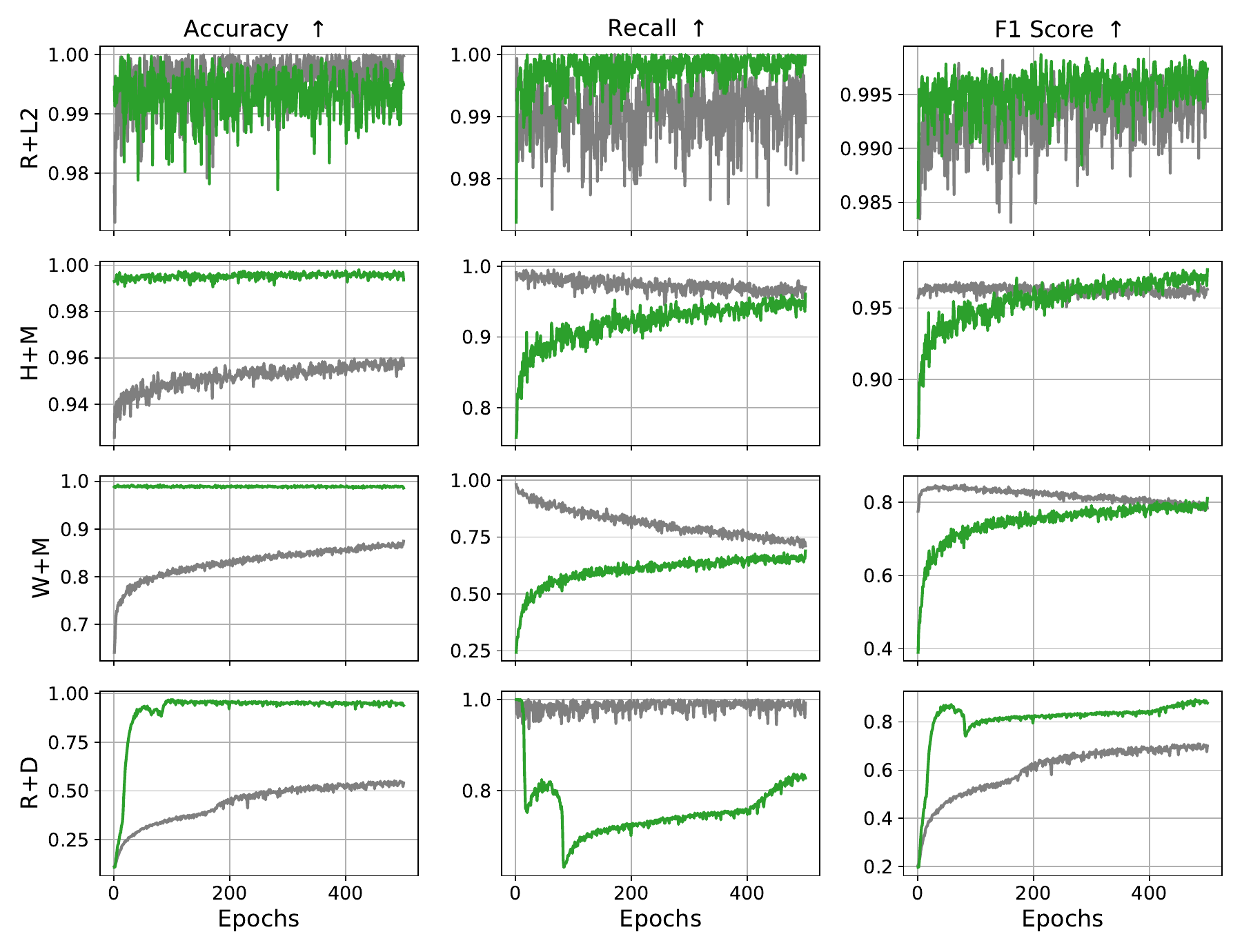}
  \caption{Comparison of the standard Flow and {\our} with Uniform Base Distribution}
  \label{fig:compare-flow}
\end{figure}

\noindent \textbf{Computational resources and runtime}: Experimented were conducted on a machine with the following specifications:
\begin{itemize}
\item GPU: NVIDIA GeForce RTX 3090 $\times$ 4
\item CPU: AMD EPYC 7402 24-Core Processor $\times$ 2
\item RAM: 968 GB
\end{itemize}

\begin{figure*}[tb]
    \centering
    \includegraphics[width=\linewidth]{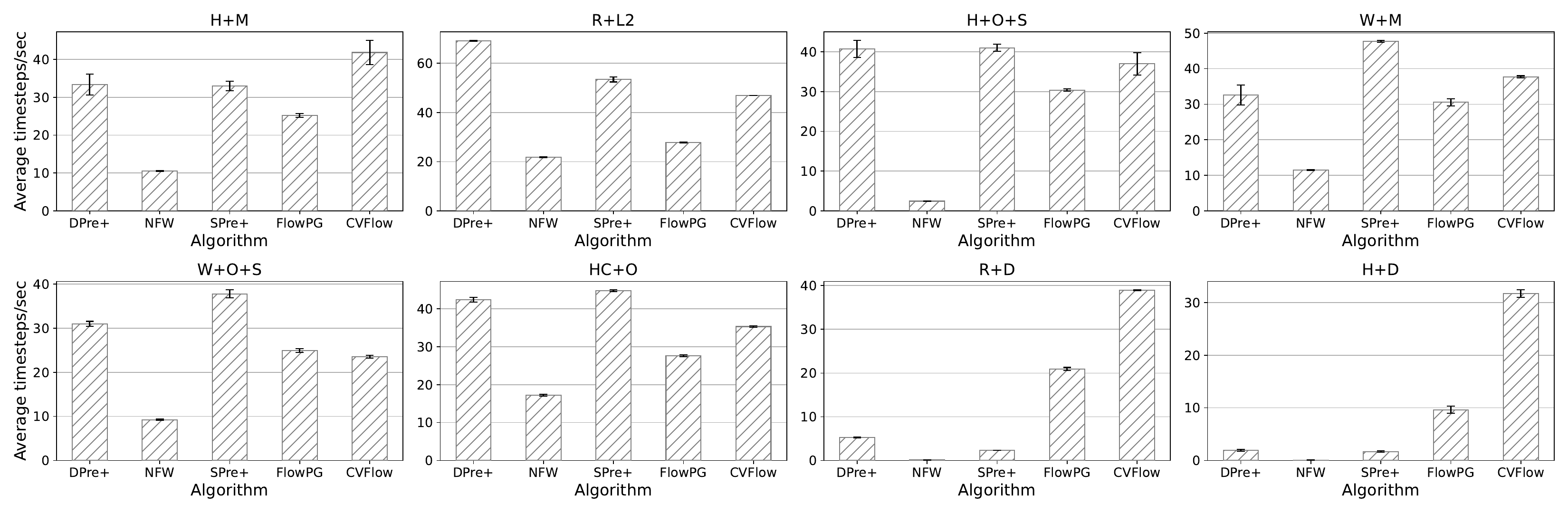}
    \caption{Average time-steps/second~$(\uparrow)$, Our approach achieves comparable speed to baselines. Specifically, with non-convex constraints (H+D, R+D), our approach significantly outpaces the baselines in terms of runtime.
    We understand the difficulty of reproducing the runtime-related results; however, we conducted each experiment on the same machine in an isolated setting to gain a better understanding of the runtime.
    }\label{sup:timesteps}
 \end{figure*}

\begin{figure*}[tb]
    \centering
    \includegraphics[width=\linewidth]{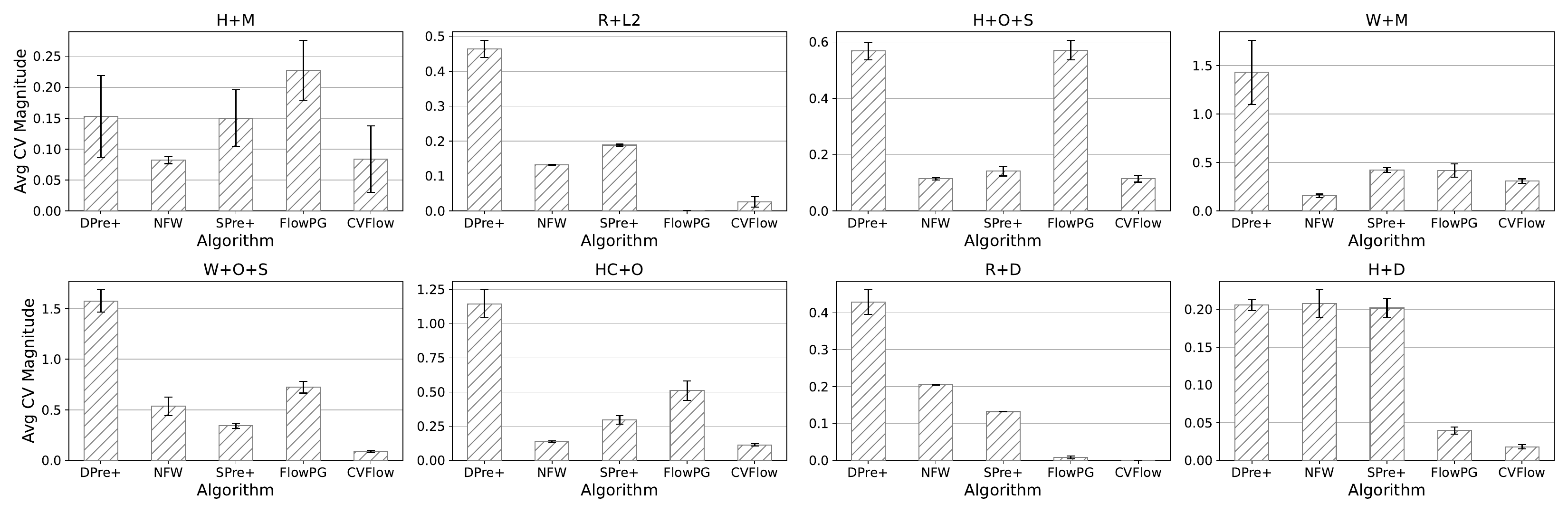}
    \caption{Average magnitude of constraint violations~$(\downarrow)$; a CV magnitude of zero implies no constraint violation. Compared to all the baselines, our approach results in a significantly lower or similar magnitude of constraint violation value, when the constraints are violated.V
    }\label{sup:cv-magnitude}
 \end{figure*}

\begin{table}[tb]
\begin{center}
\begin{tabular}{|l|ll|ll|}
\hline
 & \multicolumn{2}{l|}{CV-Count}                                     & \multicolumn{2}{l|}{Final Return}                                          \\ \hline
                              Problem & \multicolumn{1}{l|}{DDPG+CVFlow}          & SAC+CVFlow            & \multicolumn{1}{l|}{DDPG+CVFlow}               & SAC+CVFlow                \\ \hline
R+D (Non-convex)               & \multicolumn{1}{l|}{\textbf{0.01 ± 0.00}} & \textbf{0.01 ± 0.00}  & \multicolumn{1}{l|}{\textbf{22.18 ± 1.73}}     & 21.45 ± 2.46              \\ \hline
H+D (Non-convex)               & \multicolumn{1}{l|}{7.26 ± 1.97}          & \textbf{2.18 ± 1.05}  & \multicolumn{1}{l|}{3190.07 ± 101.52}          & \textbf{3236.70 ± 89.01}  \\ \hline
R+L2                           & \multicolumn{1}{l|}{\textbf{0.00 ± 0.00}} & 1.70 ± 0.22           & \multicolumn{1}{l|}{22.74 ± 1.96}              & \textbf{23.55 ± 1.95}     \\ \hline
H+M                            & \multicolumn{1}{l|}{3.26 ± 4.88}          & \textbf{0.25 ± 0.20}  & \multicolumn{1}{l|}{\textbf{3155.78 ± 207.67}} & 2935.37 ± 388.29          \\ \hline
H+O+S                          & \multicolumn{1}{l|}{2.51 ± 2.04}          & \textbf{2.42 ± 1.19}  & \multicolumn{1}{l|}{\textbf{3041.27 ± 164.10}} & 2594.44 ± 1127.36         \\ \hline
W+M                            & \multicolumn{1}{l|}{3.77 ± 2.05}          & \textbf{2.41 ± 0.72}  & \multicolumn{1}{l|}{\textbf{4263.25 ± 488.60}} & 3733.10 ± 345.92          \\ \hline
W+O+S                          & \multicolumn{1}{l|}{\textbf{1.38 ± 0.51}} & 1.55 ± 1.20           & \multicolumn{1}{l|}{\textbf{3661.52 ± 250.71}} & 3335.45 ± 200.92          \\ \hline
HC+O                           & \multicolumn{1}{l|}{34.91 ± 5.64}         & \textbf{5.04 ± 1.21}  & \multicolumn{1}{l|}{8170.54 ± 745.64}          & \textbf{9181.35 ± 680.80} \\ \hline
Ball1D                         & \multicolumn{1}{l|}{\textbf{0.00 ± 0.00}} & \textbf{0.00 ± 0.00}  & \multicolumn{1}{l|}{\textbf{62.90 ± 3.81}}     & 60.62 ± 3.38              \\ \hline
Ball3D                         & \multicolumn{1}{l|}{2.67 ± 5.87}          & \textbf{0.37 ± 0.68}  & \multicolumn{1}{l|}{\textbf{44.20 ± 3.21}}     & 41.93 ± 6.13              \\ \hline
Space-Corridor                 & \multicolumn{1}{l|}{39.08 ± 27.76}        & \textbf{12.06 ± 6.30} & \multicolumn{1}{l|}{900.00 ± 316.23}           & \textbf{1000 ± 0.00}      \\ \hline
Space-Arena                    & \multicolumn{1}{l|}{21.90 ± 14.98}        & \textbf{10.78 ± 4.07} & \multicolumn{1}{l|}{968.00 ± 65.70}            & \textbf{1000 ± 0.00}      \\ \hline
\end{tabular}
\caption{\ourcv-Flow can be integrated with DDPG as well.}
\end{center}
\end{table}

\end{document}